# The Estimation of Subjective Probabilities via Categorical Judgments of Uncertainty

Alf C. Zimmer
University of Regensburg
F.R.G.

Abstract

Theoretically as well as experimentally it is investigated how people represent their knowledge in order to make decision or to share their knowledge with others. Experiment 1 probes into the ways how people gather information about the frequencies of events and how this knowledge is interfered with by the requested response mode, that is, numerical vs. verbal estimates. The least interference occurs if the subjects are allowed to give verbal responses. From this it is concluded that processing knowledge about uncertainty categorically, that is, by means of verbal expressions, imposes less mental work load on the organism than numerical processing.

Possibility theory is used as a framework for modelling the Individual usage of verbal categories for grades of uncertainty. The 'elastic' constraints on the verbal expressions for every single subject are determined in Experiment 2 by means of sequential testing. In further experiments it is shown that the superiority of the verbal processing of knowledge about uncertainty quite generally reduces persistent biases reported in the literature: conservatism (Experiment 3) and negligence of regression (Experiment 4). In a final experiment (5) about predictions in a real-life situation it turns out that in a numerical forecasting task subjects restricted themselves to those parts of their knowledge which are numerical. On the other hand subjects in a verbal forecasting task accessed verbally as well as numerically stated knowledge.

Forcasting is structurally related to the estimation of probabilities for rare events insofar as supporting and contradicting arguments have to be evaluated and the choice of the final judgment has to be justified according to the evidence brought forward. In order to assist people in such choice situations a formal model for the interactive checking of arguments has been developed. The model transforms the normal-language quantifiers used in the arguments into fuzzy numbers and evaluates the given train of arguments by means of fuzzy numerical operations. Ambiguities in the meanings of quantifiers are resolved interactively. In order to force the subjects to argue consistently, alternative trains of arguments are presented by the model and the subjects are asked to justify their choice of an argumentative train and the rejection of alternatives.

Most investigations of subjective probabilities have been primarily concerned with the **procedures** underlying the generation of the probability judgments (see Kahneman, Slovic & Tversky, 1982 for an overview). The question how humans **represent** their knowledge about the uncertainty of events, however, has been given very little attention (except for Reyna, 1981). In most studies on subjective probability and the biases underlying these judgments it has been implicitly assumed that the information is stored symbolically (in this case in the numerical mode) and that by means of retrieval the numbers representing the knowledge about uncertainty can be given immediately and that there is no loss of information between the accessed information and the answers given. Any inconsistency between information intake (the objective side) and the numerically expressed subjective probability (the subjective side) is then ascribed to the procedures applied in retrieval. That is, the subjects has presumedly chosen an inappropriate algorithm or heuristic in either accessing the information or in deriving conclusions from it. Examples for such procedural fallacies are overconfidence, conservatism (i.e. sticking to an initial appraisal of a situation in spite of new information available for revision), and negligence of the regression effect (i.e. the implicit assumption of a perfect correlation between the predictor variable and the criterion).

The approach taken here takes off by asking how the expected frequencies of uncertain events are represented internally (e.g. (i) in a verbal propositional mode, (ii) in a numerical propositional mode, or (iii) in an analogue mode of automatic frequency monitoring). Coping with uncertainty is ubiquitous in humanking (Wright and Phillips, 1980) and verbal expressions for different degrees of certainty can be found in most languages. Except for situations like betting, people usually handle communication about uncertainty by means of verbal expressions and by the implicit or explicit rules of conversation associated with them. The research reported here therefore start from the analysis of the meaning of common verbal expressions for uncertain events. These expressions are interpreted as possibility functions (Zadeh, 1978) and the procedures applicable to them (e.g. hedging). Since this theory allows for a numerical interpretation by means of determining the elastic constraints on the usage of such expressions, the results gained by interpreting verbal expressions of uncertainty as possibility functions can be compared to the results of the above mentioned studies, where subjects had to express their judgments numerically.

The first step in the investigation of the internal representation of uncertainty was to ask how people gather the knowledge, from which by means of retrieval the verbal expressions for the probabilities of uncertain events are derived. One can assume that for repetitive events, e.g. the outcomes of ball games or the daily weather, people monitor the frequencies of outcomes automatically and revise their knowledge accordingly. Such an automatic monitoring of frequencies seems to be a plausible candidate for the initial mode of representation underlying the generation of judgments concerning subjective probabilities.

In Experiment 1 it was attempted to determine if human observers are able to monitor the frequencies of more than one unattended stimulus attribute. It turned out that the modality of the response was critical for the subjects' ability to assess frequencies of events in the unattented stimulus attributes. The results show that (i) the verbal judgments were more precise than the numerical ones, (ii) in all cases the second frequency judgments were not as precise as the first ones, but, most importantly, the impairment was less severe when the first judgment was verbal. From these results it seems plausible to conclude, first, that more than one unattended variable can be automatically monitored, but that the judgmental precision depends on the mode of probing this knowledge. Second, if more than one judgment has to be made, there is interference between them, but the amount of interference depends on the modes of the judgments. The apparent superiority of the verbal mode leads to the tentative interpretation that the verbal mode for representing knowledge processes information more effectively than the numerical mode. If the overload of the mental processing capacity necessitates the application of heuristics, and if biases in human judgment can be traced back to mistaken applications of heuristics (for an overview see Nisbett and Ross, 1980), the following conjecture seems plausible: **Any mode of judgment imposing less mental work load should be more valid than one requiring more mental processing capacity, everything else being equal.**

Modelling the meanings of verbal labels for relative frequencies is straightforward in the framework of fuzzy set theory. The universe of discourse is the unit interval and the regions of applicability, possible applicability, and inapplicability can be determined empirically, that is, the expressions are interpreted as fuzzy numbers in the unit interval. The resulting possibility functions for a given set of verbal expressions are depicted in Figure 1. The spacing of the possibility functions is reminiscent of Witte's (1960) treatment of verbal judgments stabilized in memory over time. Zimmer (1980) has shown that such equidistant and equally shaped categories are conversationally optimal, but it has to be kept in mind that this communicability constraint is a simplifying assumption, which is not a prerequisite for this kind of modelling.

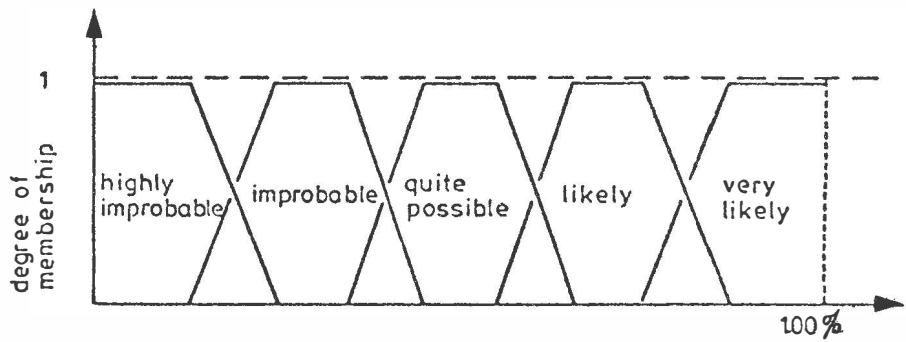

Figure 1. Possibility functions for a set of verbal expressions of uncertainty

Experiment 2 consisted of three parts. In the first part a survey was taken of the verbal expressions used by the subjects for the description of uncertain events, in part 2 the fuzzy meanings of verbal expressions of every individual subject for uncertain events were empirically determined by a modified Robbins-Monroe procedure and in part 3 these meanings were tested for calibration by comparing the individual subjective expectations of success in knowledge-test items with the actual individual probability of success as derived from the 1-parameter logistic test model.

The survey revealed that subjects differed in the number of verbal categories they spontaneously used for uncertain events. Furthermore, the meanings of these verbal categories were not the same for all subjects. In order to determine the fuzzy numbers in the unit interval which convey the meanings of the verbal expressions, subjects were asked to judge the frequencies of white dots in random dot patterns on a cr tube. The frequencies of white dots (between 5 and 95 %) were changed according to the Robbins-Monroe procedure for all verbal expressions.

The individual meanings (fuzzy numbers) of the verbal expressions were tested for calibration by giving the subjects knowledge-test items to solve and to ask how confident they were concerning the correctness of a given answer (this is a standard procedure, see Lichtenstein, Fischoff, and Phillips, 1982).

A severe problem in such calibration tasks is that individual ability and item difficulty are usually confounded because the subject's probability estimate is judged against the relative frequency of the group the subject is a member of. In order to separate the variables 'difficulty of the item' and 'ability of the subject', both of which influence the probabilities of correct answers, the conjoint-measurement approach of Rasch (1966) is applied, which provides independent estimates of the ability of a subject, $\xi_i$, and the difficulty of an item j, $\delta_j$. These estimates in turn allow for an assessment of the probability that subject i solves item J, according to the following formula:

$$p(x=1|\xi_i \& \delta_j) = \frac{e^{\xi_i - \delta_j}}{1 + e^{\xi_i - \delta_j}}$$

Figure 2 gives the difficulty-by-label curves for all subjects using

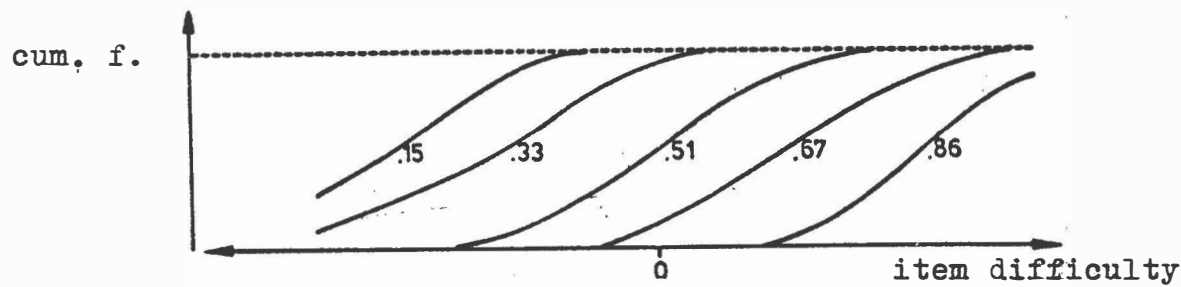

Figure 2. The average cumulative difficulty-by-label curves for all subjects applying five expressions for their subjective probabilities of successes, and the estimated numerical values

five labels (n=57). The curves are fairly parallel and the median values of the possibility functions (ranges of applicability for the expressions) are in good agreement with the probabilities estimated from the model.

A typical example for suboptimal information processing has been demonstrated in experiments on conservatism, where subjects tend to stick to their initial assumptions concerning the probability of events despite the fact that in the light of new information they should revise these assumptions. The optimal revision strategy for estimates in this task is the application of Bayes' theorem, which therefore can be used as a normative standard for the subjects' performance. In a series of experiments Phillips and Edwards (1966) have investigated this phenomenon with the result that in all cases conservatism occured. This effect could be slightly reduced by relieving the memory load of the subjects and in one condition by permitting them to answer verbally.

In a replication experiment of Phillips' & Edwards' experiment III with a ratio of 70:30 expected successes in the two bins, one group of subjects was requested to use verbal descriptions and the second group to use numerical estimates. The difference to Phillips & Edwards' procedure, however, was that the verbal as well as the numerical descriptions had been calibrated individually according to the procedure of the second experiment. That is, the responses had been interpreted as fuzzy numbers with elastic constraints.

The results of the verbal and of the numerical judgments plus those from Phillips and Edwards (1966, Experiment III) are shown in Figure 3. These results clearly indicate that verbal responses induce near optimal, or Bayesian, performance in the subjects. In contrast, the calibrated numerical judgment are only slightly better than the results of Phillips & Edwards. In the light of this result one can assume that the so called conservatism in information processing is not due to an improper processing of the information provided but that it is due to the difficulties subjects have in storing and processing numerical expressions. Thus, it is not the perception of reality that is biased, but it is the mode of expressing one's knowledge about reality.

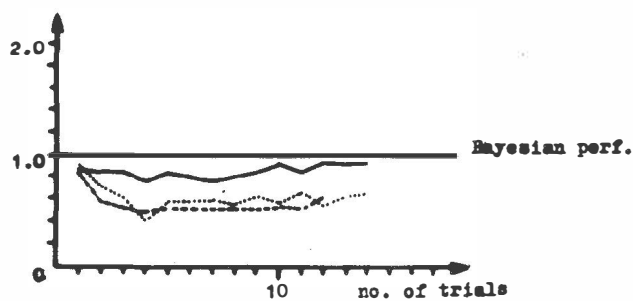

Figure 3. Comparison of Phillips & Edwards results (broken line) with the results from individually calibrated verbal responses (unbroken line) and individually calibrated numerical responses (dotted line).

A second prototypical example for people's difficulties in handling uncertainty can be found in the prediction of uncertain events, Kahneman & Tversky (1973) studied categorical as well as numerical predictions and found that in all cases subjects tended to neglect the base rates and exhibited mistaken intuitions about regression For instance, they acted as if in all cases the correlation between the predictor variable and the criterion was perfect and as if the predictor variable was measured with perfect reliability.

In Experiment 4 subjects were asked to predict the individual success of students at the university from their performance in highschool. In order to let subjects develop their notion of the correlational relationship between highschool and college performance, they were given a sample of 75 typical cases of grades students received at highschool and in the graduating exam at the university. Afterwards they had to predict the performance at the university of 50 students on the basis of their highschool grades. The results were very similar to those obtained by Kahneman and Tversky (1973) but when the subjects were asked how certain they were about the correctness of their predictions, they gave quite low subjective degrees of confidence.

When probed further about the direction of their probable error of prediction 27 out of 30 subjects indicated correctly that the true value would probably be closer to the mean performance than they had predicted.

This result as well as those reported above seem to indicate that subjects are better able to take into account complex dependencies by means of verbal processing than if they are forced to process the same amount of information numerically. As Gregory (1982) points out, numbers and computation form a more recent tool of mind than language and therefore the numerical information processing is less automatic.

In experiment 5 subjects (24 bank clerks responsible for foreign exchange) were asked to predict what the exchange rate between the US Dollar and the Deutschmark would be four weeks later. Twelve subjects had to give the predictions "in their own words as they would talk to a client", whereas the other twelve were asked to give numerical estimates in percentage of change. Both groups were asked to verbalize the steps they took in order to come up with the prediction. After the predictions of the first group had been calibrated with a technique similar to that used in Experiment 2, they were compared to the predictions made by the numerical forecasting group. It turned out that the first group was more correct and more internally consistent. While this is interesting in itself, another point is more important: The slight difference in the instructions caused marked differences in the way the subjects performed their task as revealed by the verbal protocol. The verbal prediction group used quantitative variables (e.g. the GNP increase in percent) as well as qualitative variables (e.g. the stability of the German government) for deriving their predictions, whereas the other group merely took into account those variables which are usually expressed numerically. From this it seems plausible to assume that one reason for the superiority in the verbal forecasting condition is the fact that the knowledge base on which these subjects relied was broader and allowed for more elaboration. However, it has to be kept in mind that the heuristic of causal schemata can be also misleading; Nisbett & Ross (1980) report ample evidence for the deleterious effects of misinterpreting diagnostic information as causal. The major difference between the studies reported in Nisbett & Ross (1980) and this experiment lies in the fact that the bank clerks were actively searching for information and only implemented their own knowledge into their reasoning.

The studies of Begg (1982), Zimmer (1982, in press), as well as the theoretical analysis of "rational belief" by Kyburg (1983) indicate that describing human reasoning in the framework of classical logic might be a mistaken approach. The proposed alternative starts from the assumption that people usually start with making a claim about a given problem (e.g. the estimation of the probability of a rare event). Afterwards they justify this claim by giving the underlying train of arguments and the available evidence favorable for the claim. Counter-arguments and/or contracting evidence forces them either to revise the claim or to refute the argumentative alternatives.

We are developing a model which helps the decision maker to check the arguments by which he or she backs or justifies the claim made. A special kind of justifications are those with explicit (e.g. "Often X is influenced by Y") or implicit quantification (e.g. "Xs prevail if Y"). If the meaning of the quantifier ar unambiguous, it is represented as a fuzzy number in the interval [0, 100 %] giving the expected proportion of positive incidences of X (see Zadeh, 1984; Zimmer, 1984). In the case of ambiguous quantifiers (e.g. "usually") the meaning is interactively rendered so precise that the quantifier can be expressed as a single fuzzy number (see Figure 4).

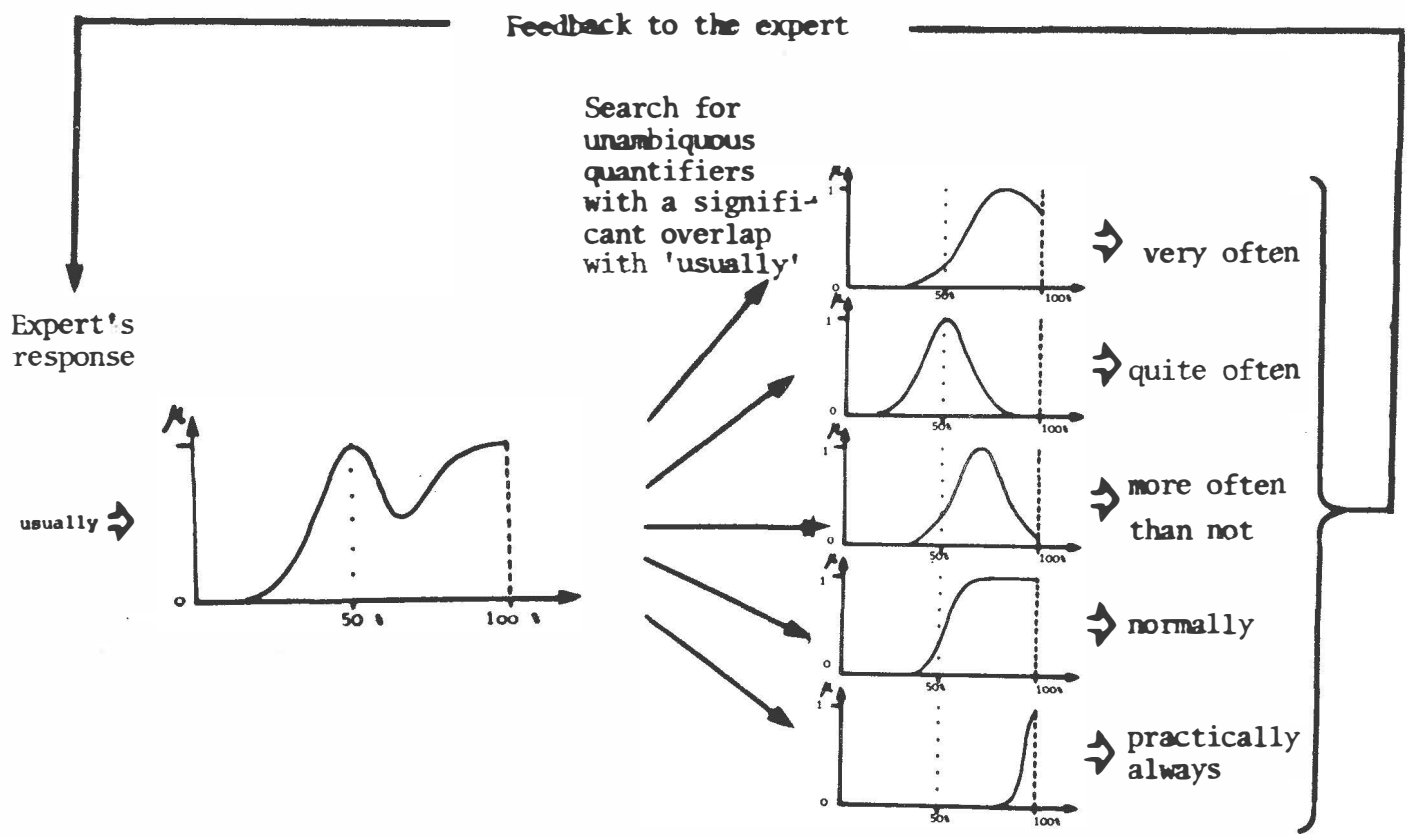

Figure 4. Interactive determination of the meaning of an intended quantifier ( indicates translation).

For a final evaluation these local interpretations of argumentative propositions are propagated and it is checked how and if the reasons given are backed by factual knowledge (e.g. "On which observable instances do you base the statement 'X'?") and how reliable and generalizable these supporting facts are. After all the components of the argument have been elicited and after the meaning of the used predicates and the credibility of the ground, the warrant, the backing, and of rebuttals (Toulmin, Rieke & Janik, 1979) have been determined, the credibility of the claim is analytically derived from these values by means of fuzzy syllogistic reasoning (Zadeh, 1984; Zimmer, 1984). The credibility values are either transformed into verbal expressions (e.g. 'likely', 'unprobable' etc., see Zimmer, 1983) or are given in their original numerical form together with the elastic constraints. If the subjective evaluation of the claim by the decision maker and the analytical evaluation are about the same, the interactive process ends. If, however, the expert disagrees, he or she is asked to give further grounds or to revise the credibility ratings for the facts given.

The interactive model for the elucidation of arguments underlying the claims (e.g. predictions, diagnoses) of experts on the on hand serves as a means for an unbiased probability assessment for claims. Insofar it resembles the procedure proposed by Henrion and Morgan (see Morgan, in press). On the other hand, however, it makes explicit the knowledge base on which the expert grounds his/her claim. The comparison of the knowledge bases underlying the predictions of different experts for the same event shows if these predictions are based on more or less same reasons or not. In the first case an agglommeration of the evaluations made by different experts is admissible. In the other case, however, only those judgments can be pooled which are based on comparable knowledge bases.